\documentclass{article}
\PassOptionsToPackage{square,comma,numbers,compress}{natbib}
\usepackage[preprint]{preprint}

\usepackage[utf8]{inputenc} %
\usepackage[T1]{fontenc}    %
\usepackage[colorlinks,bookmarks=false]{hyperref}%
\usepackage{url}            %
\usepackage{booktabs}       %
\usepackage{amsfonts}       %
\usepackage{nicefrac}       %
\usepackage{microtype}      %
\usepackage[table]{xcolor}
\usepackage{tikz}

\usepackage{booktabs}
\usepackage{multicol}
\usepackage{multirow}
\usepackage{color}
\usepackage{caption}
\usepackage{subcaption}
\usepackage{hhline}
\usepackage{pifont}
\usepackage{threeparttable}
\usepackage{makecell}
\usepackage{algorithm} 
\usepackage{listings}
\usepackage{amsfonts,amssymb}
\usepackage{amsmath}
\usepackage{graphicx}
\usepackage{lipsum}
\usepackage{arydshln}
\usepackage[accsupp]{axessibility}
\usepackage{enumitem}
\usepackage{tcolorbox}

\newcommand{\benchnametwo}{MM-Vet v2}

\def\eg{\emph{e.g.}}
\def\etc{\emph{etc}}

\newlength\savedwidth
\newcommand\whline{\noalign{\global\savedwidth\arrayrulewidth\global\arrayrulewidth 0.8pt}\hline\noalign{\global\arrayrulewidth\savedwidth}}
\usepackage{verbatim}

\newcommand{\greentext}[1]{\textcolor[rgb]{0, 0.5, 0}{#1}}
\newcommand{\orangetext}[1]{\textcolor[RGB]{255, 69, 0}{#1}}
\newcommand{\bluetext}[1]{\textcolor[RGB]{0, 0, 255}{#1}}
\newcommand{\purpletext}[1]{\textcolor[RGB]{148, 0, 211}{#1}}

\newcommand{\firstc}{{\cellcolor[RGB]{ 171, 235, 198 }}}
\newcommand{\secondc}{{\cellcolor[RGB]{ 253, 235, 208}}}
\newcommand{\thirdc}{{\cellcolor[RGB]{  214, 234, 248  }}}

\newcommand*\samethanks[1][\value{footnote}]{\footnotemark[#1]}

\title{
\benchnametwo{}: A Challenging Benchmark to Evaluate Large Multimodal Models for Integrated Capabilities}

\author{Weihao Yu$^1$\thanks{Equal contribution.}~~~~Zhengyuan Yang$^2$\samethanks~~~~Lingfeng Ren$^1$\samethanks~~~~Linjie Li$^2$~~~Jianfeng Wang$^2$ \\ \textbf{Kevin Lin$^2$~~~Chung-Ching Lin$^2$~~~
Zicheng Liu$^3$~~~Lijuan Wang$^2$\thanks{Corresponding authors.}~~~~Xinchao Wang$^1$\samethanks} \\
$^1$National University of Singapore \qquad $^2$Microsoft \qquad $^3$Advanced Micro Devices\\ 
{\tt\footnotesize \{weihaoyu,lingfengren\}@u.nus.edu \quad zicliu@outlook.com \quad xinchao@nus.edu.sg } \\
{\tt\footnotesize \{zhengyang,lindsey.li,jianfw,keli,chungching.lin,lijuanw\}@microsoft.com}\\
\\
\makebox[\textwidth][l]{\small{\includegraphics[height=1em]{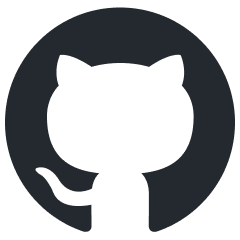} Code \& data: \url{https://github.com/yuweihao/MM-Vet}}} \\
\makebox[\textwidth][l]{\small{\includegraphics[height=1em]{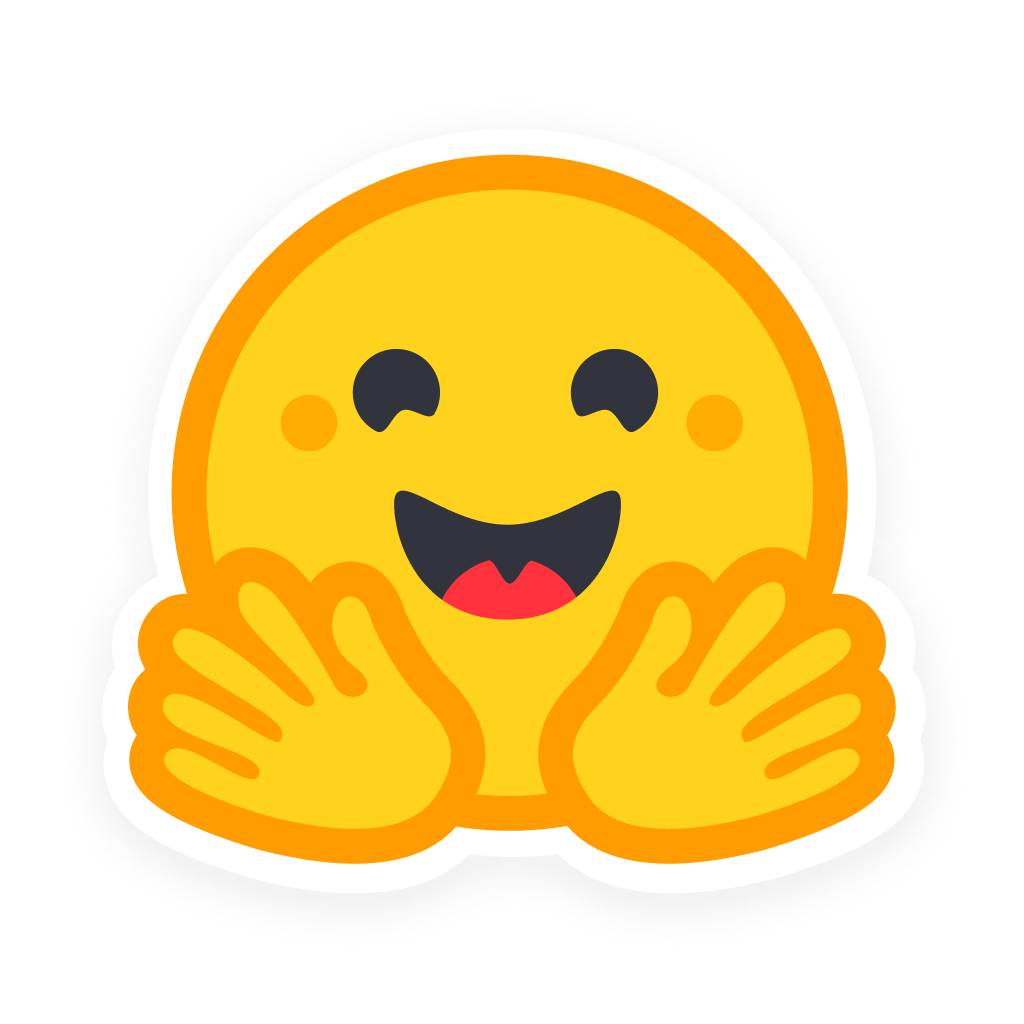} Online evaluator: \url{https://huggingface.co/spaces/whyu/MM-Vet-v2_Evaluator}}} \\
\makebox[\textwidth][l]{\small{\includegraphics[height=1em]{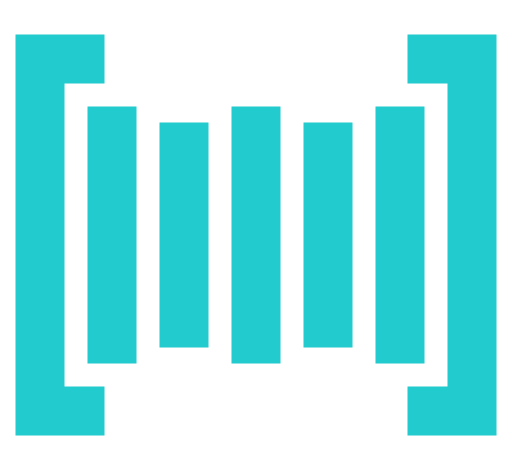} Leaderboard: \url{https://paperswithcode.com/sota/visual-question-answering-on-mm-vet-v2}}}
}

\begin{document}

\maketitle

\begin{tikzpicture}[remember picture,overlay,shift={(current page.north west)}]
\node[anchor=north west,xshift=0.7cm,yshift=-2cm]{\scalebox{-1}[1]{\includegraphics[width=3cm]{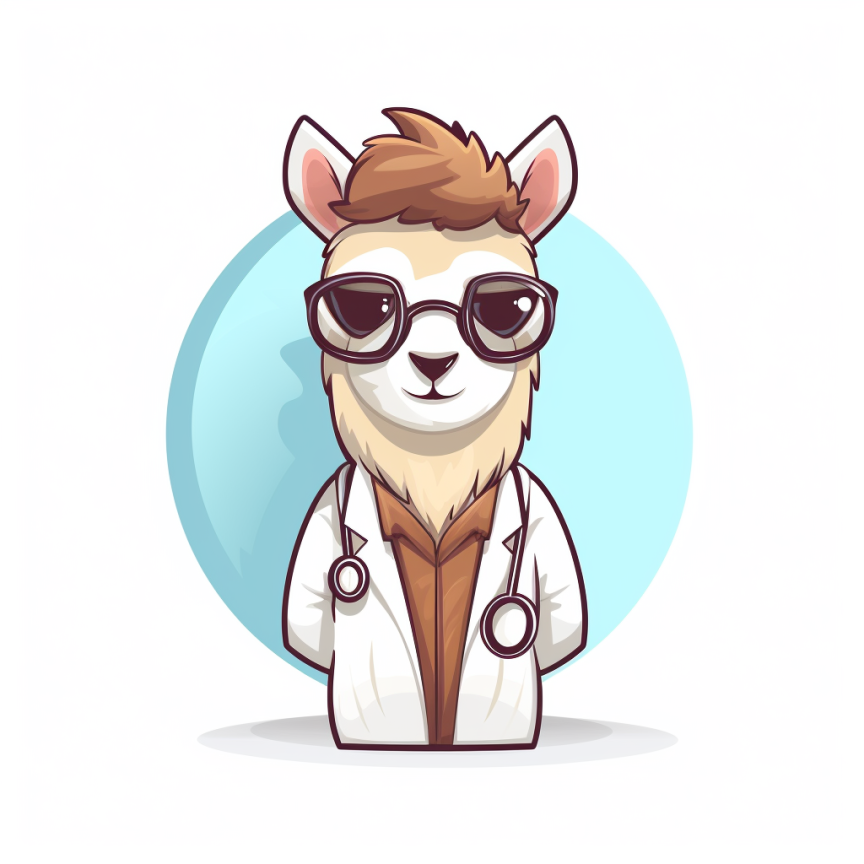}}};
\end{tikzpicture}

\vspace{-25pt}
\begin{abstract}
MM-Vet, with open-ended vision-language questions targeting at evaluating integrated capabilities, has become one of the most popular benchmarks for large multimodal model evaluation. MM-Vet assesses six core vision-language (VL) capabilities: recognition, knowledge, spatial awareness, language generation, OCR, and math. However, its question format is restricted to single image-text pairs, lacking the interleaved image and text sequences prevalent in real-world scenarios. To address this limitation, we introduce MM-Vet v2, which includes a new VL capability called ``image-text sequence understanding'', evaluating models' ability to process VL sequences. Furthermore, we maintain the high quality of evaluation samples while further expanding the evaluation set size. Using MM-Vet v2 to benchmark large multimodal models, we found that Claude 3.5 Sonnet is the best model with a score of 71.8, slightly outperforming GPT-4o which scored 71.0. Among open-weight models, InternVL2-Llama3-76B leads with a score of 68.4.

\end{abstract}

\section{Introduction}

\begin{figure}[htbp]
  \centering
   \includegraphics[width=0.9\linewidth]{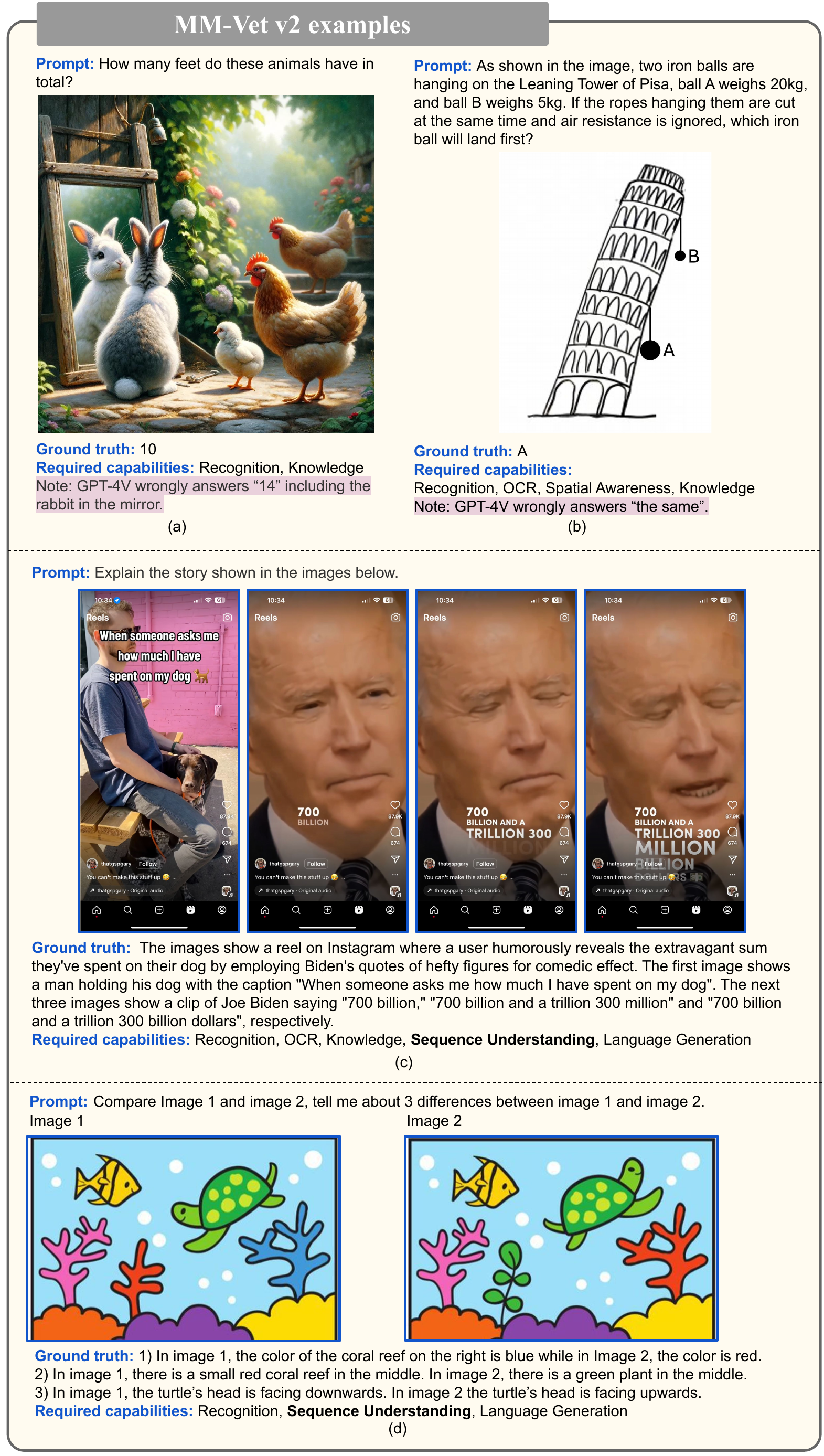}
   \caption{Four examples from \benchnametwo{}. Compared with MM-Vet \cite{yu2023mm}, \benchnametwo{} introduces more high-quality evaluation samples (\eg, (a) and (b)), and the ones with the new capability of image-text sequence understanding (\eg, (c) and (d)).
   }
   \label{fig:first_figure}
\end{figure}

Large multimodal models (LMMs)~\cite{llava,gpt4v,alayrac2022flamingo} evolve rapidly, demonstrating emergent abilities to solve complex tasks~\cite{yang2023dawn} that required multiple integrated capabilities, such as GUI navigation~\cite{yan2023gpt,you2024ferret,qinghong2024videogui}, screenshot to code~\cite{si2024design2code,han2023chartllama}, video understanding~\cite{lin2023mm,zhang2024mm}, \etc. To comprehensively evaluate LMMs, multiple benchmarks have been proposed, such as MME~\cite{fu2023mme}, MMBench~\cite{liu2023mmbench}, SEED-Bench~\cite{li2023seedbench}, MMMU~\cite{yue2024mmmu}, and MM-Vet~\cite{yu2023mm}. Notably, MM-Vet is designed to evaluate LMMs from a capability integration perspective, defining tasks based on their required core capabilities. The benchmark accepts open-ended responses and takes GPT-4~\cite{openai2023gpt4} to score model predictions, which better align with real-world scenarios. Such effective evaluation designs make MM-Vet widely utilized as a standard benchmark for LMM evaluation, as indicated by its leaderboard \footnote{MM-Vet leaderboard: \url{https://paperswithcode.com/sota/visual-question-answering-on-mm-vet}}.%

Despite its popularity, MM-Vet and other concurrent evaluation benchmarks~\cite{fu2023mme,liu2023mmbench,li2023seedbench} are beginning to fall short in assessing the more advanced capabilities that have emerged in the latest LMMs, such as GPT-4V~\cite{gpt4v} and its successors~\cite{sun2023generative,internvl,llava,qwenvl,gemini1_5,claude3}. Specifically, one major limitation is the question format. The questions in MM-Vet are limited to a single image-text pair, lacking the capability to handle interleaved image and text sequences. This design choice was natural at the time of prior studies~\cite{yu2023mm}, given that most LMMs only supported single image inputs. However, the ability to process arbitrarily interleaved image-text sequences is crucial for advanced LMMs and should be included in LMM evaluation. %

In addition to the six core capabilities defined in MM-Vet, we introduce an additional capability: ``image-text sequence understanding.'' This measures the LMMs' ability to process image-text sequential data, as shown in Figure \ref{fig:first_figure}. For example, to complete the task in Figure \ref{fig:first_figure}(c), the LMM needs to understand the text question followed by multiple sequential images. This task requires image-text sequence understanding alongside the capabilities of recognition, OCR, knowledge, and language generation defined in MM-Vet. The image-text sequence can contain multiple images of drastically different usages, such as multiple video frames in Figure \ref{fig:first_figure}(c) for temporal understanding, and spot the difference challenge in Figure \ref{fig:first_figure}(d) for image comparison. These capabilities to comprehend arbitrarily interleaved image-text sequences is one fundamental step towards stronger general intelligence.

Another limitation is the size of the evaluation set. MM-Vet only has 217 evaluation samples due to the difficulty of high-quality data collection. We aim to maintain the same high quality while further expanding the evaluation set size. 
We break down the challenge of creating high-quality evaluation samples into two steps: generating high-quality questions and producing reference answers. For the question generation, we find it difficult for crowd-sourcing workers to propose a meaningful and complicated question that covers various scenarios, \eg, in Figure~\ref{fig:first_figure} (a,b). 
Instead, we have researchers design and collect 517 questions covering various scenarios from daily life to expert/industry applications, extended from an exploratory report on GPT-4V~\cite{yang2023dawn}. For straightforward questions that can be answered in a few words, our experts directly annotate the reference answer. For questions that need long paragraphs to answer, we first employ the GPT-4V \cite{gpt4v} to draft the response. Next, our experts correct draft if there is any error and then rephrase it into the final ground truth.

After developing MM-Vet v2, we evaluate multiple advanced LMMs. Claude 3.5 Sonnet \cite{claude3.5sonnet_blog} achieves the highest performance with a score of 71.8, slightly surpassing GPT-4o \cite{gpt4oblog} by 0.8 points. Notably, InternVL2-Llama3-76B \cite{internvl2blog}, an open-weight model, also delivers a very competitive score of 68.4.

\begin{figure}[t]
  \centering
   \includegraphics[width=0.9\linewidth]{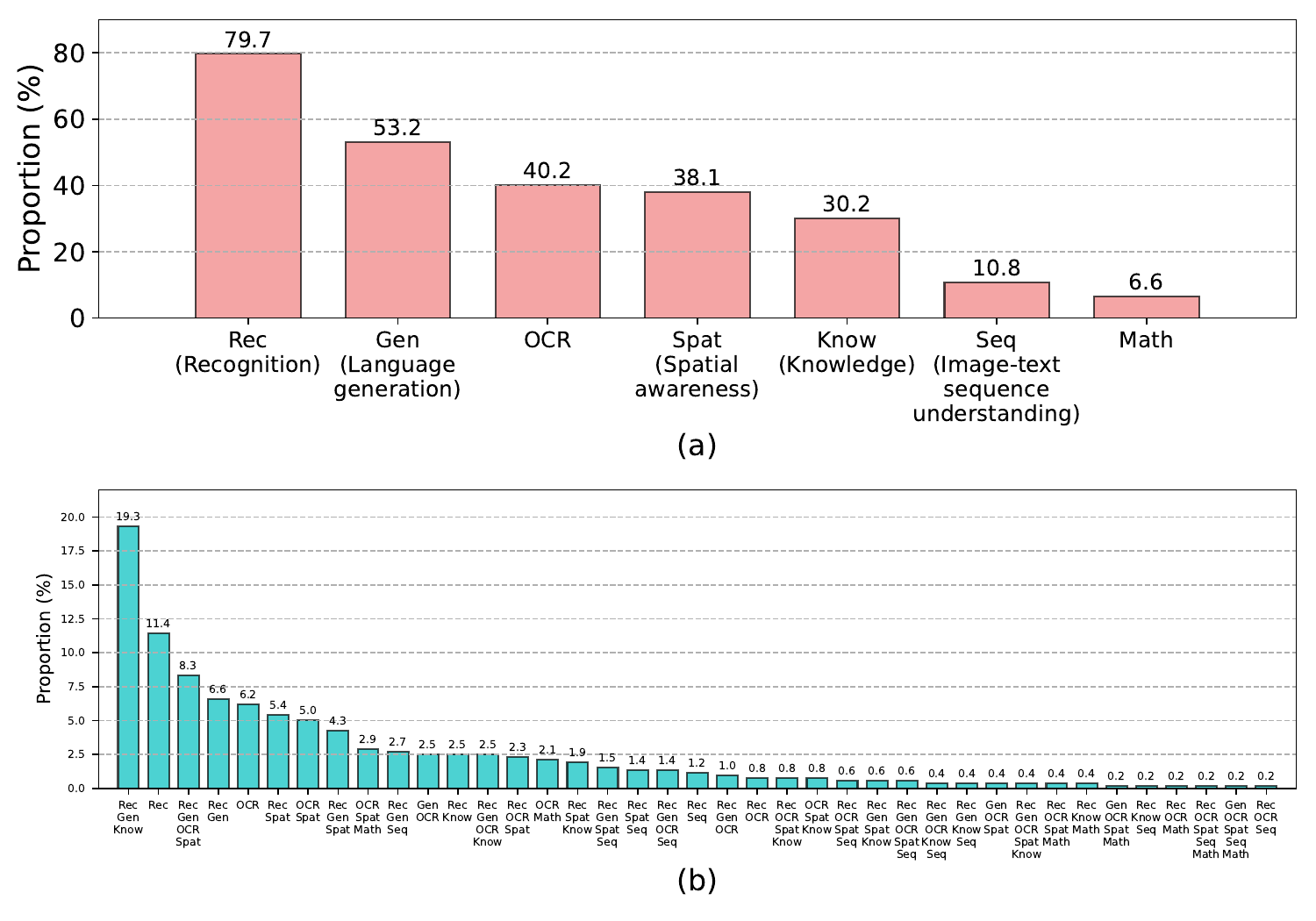}
   \vspace{-10pt}
   \caption{Proportions of (a) core capabilities and (b) capability integrations on \benchnametwo{}.
   }
   \label{fig:dataset_cap}
\end{figure}

\section{Dataset and evaluator}

\begin{table}[t]
\caption{\benchnametwo{}'s few-shot prompt to evaluate model outputs using GPT-4 (gpt-4-0613), where \bluetext{$\mathcal{Q}$} represents a sample's question, \greentext{$\mathcal{G}$} denotes the ground truth and \greentext{$\mathcal{P}$} is the model output for the sample. Compared with MM-Vet, MM-Vet v2 adds <image> to represent image position in the question. 
}
\label{tab:v2prompt}
\centering
\begin{tcolorbox} 
    \centering
    \small
    \begin{tabular}{p{0.985\columnwidth}}

Compare the ground truth and prediction from AI models, to give a correctness score for the prediction. <image> in the question indicates where an image is. <AND> in the ground truth means it is totally right only when all elements in the ground truth are present in the prediction, and <OR> means it is totally right when any one element in the ground truth is present in the prediction. The correctness score is 0.0 (totally wrong), 0.1, 0.2, 0.3, 0.4, 0.5, 0.6, 0.7, 0.8, 0.9, or 1.0 (totally right). Just complete the last space of the correctness score. \\
\\
| \bluetext{Question} | \greentext{Ground truth} | \orangetext{Prediction} | \purpletext{Correctness} | \\
| --- | --- | --- | --- | \\
| \bluetext{What is x in the equation?<image>} | \greentext{-1 <AND> -5} | \orangetext{x = 3} | \purpletext{0.0} | \\
| \bluetext{What is x in the equation?<image>} | \greentext{-1 <AND> -5} | \orangetext{x = -1} | \purpletext{0.5} | \\
| \bluetext{What is x in the equation?<image>} | \greentext{-1 <AND> -5} | \orangetext{x = -5} | \purpletext{0.5} | \\
| \bluetext{What is x in the equation?<image>} | \greentext{-1 <AND> -5} | \orangetext{x = -5 or 5} | \purpletext{0.5} | \\
| \bluetext{What is x in the equation?<image>} | \greentext{-1 <AND> -5} | \orangetext{x = -1 or x = -5} | \purpletext{1.0} | \\
| \bluetext{Can you explain this meme?<image>} | \greentext{This meme is poking fun at the fact that the names of the countries Iceland and Greenland are misleading. Despite its name, Iceland is known for its beautiful green landscapes, while Greenland is mostly covered in ice and snow. The meme is saying that the person has trust issues because the names of these countries do not accurately represent their landscapes.} | \orangetext{The meme talks about Iceland and Greenland. It's pointing out that despite their names, Iceland is not very icy and Greenland isn't very green.} | \purpletext{0.4} | \\
| \bluetext{Can you explain this meme?<image>} | \greentext{This meme is poking fun at the fact that the names of the countries Iceland and Greenland are misleading. Despite its name, Iceland is known for its beautiful green landscapes, while Greenland is mostly covered in ice and snow. The meme is saying that the person has trust issues because the names of these countries do not accurately represent their landscapes.} | \orangetext{The meme is using humor to point out the misleading nature of Iceland's and Greenland's names. Iceland, despite its name, has lush green landscapes while Greenland is mostly covered in ice and snow. The text `This is why I have trust issues' is a playful way to suggest that these contradictions can lead to distrust or confusion. The humor in this meme is derived from the unexpected contrast between the names of the countries and their actual physical characteristics.} | \purpletext{1.0} | \\
| \bluetext{$\mathcal{Q}$} | \greentext{$\mathcal{G}$} | \greentext{$\mathcal{P}$} |

    \end{tabular}
\end{tcolorbox}

\end{table}

The same as MM-Vet \cite{yu2023mm}, we aim to build a high-quality evaluation set for large multimodal models. MM-Vet \cite{yu2023mm} defines six core vision-language capabilities, including recognition (Rec), knowledge (Know), OCR, spatial awareness (Spat), language generation (Gen), and Math. MM-Vet's question format is only an image-text pair, which obviously cannot measure the capability of processing sequential image and text data. To fill this gap, we introduce a new capability:
\begin{itemize}
    \item \textbf{Image-text sequence understanding (Seq).} It refers to the capability to understand and reason the relationships among sequential image and text streaming data. 
\end{itemize}

Then we carefully design 517 questions that require one or more capabilities, extended from MM-Vet \cite{yu2023mm} and an exploratory report based on \cite{yang2023dawn}. Specifically, 218 questions are directly from MM-Vet \cite{yu2023mm}, and the others are newly collected in various domains from daily life to expert/industry applications. For those questions that can be answered shortly, we annotated the ground truths directly. For questions requiring long text to respond, we first utilize GPT-4V \cite{gpt4v} to generate the answer drafts, then we carefully correct the wrong information in the drafts and rephrase them into final ground truths. Because this new dataset is the extension of MM-Vet \cite{yu2023mm}, we name it \benchnametwo{}. See Figure \ref{fig:dataset_cap} for the proportions of each core capability and capability integration.

Next, we design the evaluator for \benchnametwo{}, based on that of MM-Vet \cite{yu2023mm}. Different from MM-Vet, the input format of \benchnametwo{} is not only an image-text pair, it can also be image-text sequences. To demonstrate the position of the image in the sequences, we utilize <image> to represent the image position. The few-shot prompt for \benchnametwo{} is shown in Table \ref{tab:v2prompt}.

\section{Experiments}

\begin{table}[t]
  \caption{MM-Vet v2 evaluation results on various LMMs regarding each \textit{core VL capability}. For each column, the highest, the second, and the third highest figures are highlighted by \setlength{\fboxsep}{0pt}\colorbox[RGB]{ 171, 235, 198 }{green}, \setlength{\fboxsep}{0pt}\colorbox[RGB]{ 253, 235, 208}{orange} and \setlength{\fboxsep}{0pt}\colorbox[RGB]{  214, 234, 248  }{blue} backgrounds. All the numbers are presented in \% and the full score is 100\%.
  }
  \label{tab:results_cap}
  \centering
    \small
    \setlength{\tabcolsep}{5.3pt}
    \begin{tabular}{l|c|cccccccccc}
        \whline
        Model & MM-Vet & Rec & Gen & OCR & Spat & Know & Seq & Math & MM-Vet-v2  \\ 
        \whline
        OpenFlamingo-9B \cite{alayrac2022flamingo, awadalla2023openflamingo} & 24.8 & 19.1 & 11.0 & 11.2 & 13.0 & 18.4 & 10.2 & 2.9 & 17.6$\pm$0.2 \\
        Otter-9B \cite{li2023otter} & 24.7 & 25.1 & 16.5 & 13.7 & 19.5 & 23.5 & 9.9 & 6.4 & 23.2$\pm$0.1 \\
        LLaVA-v1.5-7B \cite{llava} & 34.2 & 30.4 & 21.5 & 21.7 & 25.4 & 23.6 & 9.5 & 6.0 & 28.3$\pm$0.2\\
        LLaVA-v1.5-13B \cite{llava} & 39.2 & 34.8 & 29.2 & 28.7 & 29.1 & 29.4 & 17.8 & 8.8 & 33.2$\pm$0.1\\
        CogAgent-Chat \cite{cogagent} & 40.5 & 33.6 & 29.2 & 35.3 & 27.3 & 33.9 & 24.7 & 5.9 & 34.7$\pm$0.2 \\
        Emu2-Chat \cite{emu2} & 45.5 & 37.9 & 28.8 & 35.1 & 35.1 & 37.9 & 18.7 & 10.8 & 38.0$\pm$0.1 \\
        IXC2-VL-7B \cite{ixc2} & 52.3 & 40.7 & 37.0 & 42.5 & 37.0 & 39.3 & 5.2 & 36.1 & 42.5$\pm$0.3\\
        CogVLM-Chat \cite{cogvlm} & 52.6 & 46.3 & 40.7 & 41.6 & 40.3 & 47.5 & 33.8 & 7.7 & 45.1$\pm$0.2\\
        InternVL-Chat-V1-2 \cite{internvl} & 49.7 & 46.1 & 40.8 & 43.8 & 41.4 & 42.1 & 25.1 & 11.8 & 45.5$\pm$0.1\\
        LLaVA-NeXT-34B \cite{llava} & 60.2 & 49.3 & 48.9 & 52.8 & 48.3 & 49.9 & 18.5 & 35.1 & 50.9$\pm$0.1\\
        InternVL-Chat-V1-5 \cite{internvl} & 62.8 & 52.0 & 48.9 & 51.0 & 49.3 & 48.2 & 37.6 & 16.5 & 51.5$\pm$0.2 \\
        InternVL2-40B \cite{internvl2blog} & 61.8 & 63.6 & 63.9 & 64.3 & 60.0 & 60.9 & 48.5 & 45.3 & 63.8$\pm$0.2 \\
        InternVL2-Llama3-76B \cite{internvl2blog} & 64.4 & \thirdc{} 67.0 & \thirdc{} 68.5 & 70.6 & 65.2 & \thirdc{} 62.4 & 59.4 & 55.4 & \thirdc{} 68.4$\pm$0.3 \\
        \hdashline 
        Claude 3 Opus \cite{claude3} & 58.6 & 53.5 & 57.6 & 60.7 & 50.0 & 51.3 & 46.1 & 42.9 & 55.8$\pm$0.2 \\
        Qwen-VL-Max \cite{qwenvl} & 66.6 & 51.7 & 51.1 & 60.2 & 49.0 & 52.5 & 27.3 & 60.2 & 55.8$\pm$0.2 \\
        Gemini Pro Vision \cite{gemini} & 63.1 & 54.3 & 50.8 & 61.5 & 55.8 & 50.4 & 45.4 & 43.5 & 57.2$\pm$0.2 \\
        GPT-4V \cite{gpt4v} & \thirdc{} 67.7 & 63.1 & 67.1 & 73.6 & 65.8 & 53.4 & 62.2 & 68.3 & 66.3$\pm$0.2 \\
        Gemini 1.5 Pro \cite{gemini1_5} & 65.8 & 64.4 & 64.7 & \thirdc{} 75.1 & \secondc{} 65.9 & 56.6 & \thirdc{} 63.9 & \thirdc{} 61.5 & 66.9$\pm$0.2 \\
        GPT-4o \cite{gpt4oblog} & \secondc{} 69.3 & \secondc{} 67.5 & \secondc{} 70.5 & \secondc{} 78.0 & \firstc{} 68.8 & \secondc{} 63.8 & \firstc{} 74.3 & \firstc{} 77.6 & \secondc{} 71.0$\pm$0.2 \\
        Claude 3.5 Sonnet \cite{claude3.5sonnet_blog} & \firstc{} 74.2 & \firstc{} 69.2 & \firstc{} 70.8 & \firstc{} 78.9 & \firstc{} 68.8 & \firstc{} 65.6 & \secondc{} 67.7 & \secondc{} 69.1 & \firstc{} 71.8$\pm$0.2 \\
        \whline
    \end{tabular}
\end{table}

\begin{table}[h!]
  \footnotesize
  \setlength{\tabcolsep}{2pt}
  
  \caption{\benchnametwo{} evaluation results on various LMMs regarding \textit{capability integrations}. Due to space limitations, only the 16 integrations with the highest proportions are displayed.
  For each column, the highest, the second, and the third highest figures are highlighted by \setlength{\fboxsep}{0pt}\colorbox[RGB]{ 171, 235, 198 }{green}, \setlength{\fboxsep}{0pt}\colorbox[RGB]{ 253, 235, 208}{orange} and \setlength{\fboxsep}{0pt}\colorbox[RGB]{  214, 234, 248  }{blue} backgrounds. All the numbers are presented in \% and the full score is 100\%.
  }
  \label{tab:results_cap_integration}
  \centering
    \scalebox{0.78}{
    \begin{tabular}{lcccccccccccccccccc}
    \toprule
        \multirow{4}{*}{\makecell[c]{Model}} & \multirow{4}{*}{\makecell[c]{~ \\ Rec \\ Gen \\ Know}} & \multirow{4}{*}{\makecell[c]{~ \\ ~ \\ ~ \\ Rec }} & \multirow{4}{*}{\makecell[c]{Rec \\ Gen \\ OCR \\ Spat}} & \multirow{4}{*}{\makecell[c]{~ \\ ~ \\ Rec \\ Gen}} & \multirow{4}{*}{\makecell[c]{~ \\ ~ \\ ~ \\ OCR}} & \multirow{4}{*}{\makecell[c]{~ \\ ~ \\ Rec \\ Spat}} & \multirow{4}{*}{\makecell[c]{~ \\ ~ \\ OCR \\ Spat}} & \multirow{4}{*}{\makecell[c]{~ \\ Rec \\ Gen \\ Spat}} & \multirow{4}{*}{\makecell[c]{~ \\ OCR \\ Spat \\ Math }}  & \multirow{4}{*}{\makecell[c]{~ \\ Rec \\ Gen \\ Seq }} & \multirow{4}{*}{\makecell[c]{~ \\ ~ \\ Gen \\ OCR }} & \multirow{4}{*}{\makecell[c]{~ \\ ~ \\ Rec \\ Know }} & \multirow{4}{*}{\makecell[c]{Rec \\ Gen \\ OCR \\ Know }} & \multirow{4}{*}{\makecell[c]{~ \\ Rec \\ OCR \\ Spat}}  & \multirow{4}{*}{\makecell[c]{~ \\ ~ \\ OCR \\ Math }} & \multirow{4}{*}{\makecell[c]{~ \\ Rec \\ Spat \\ Know}} & \multirow{4}{*}{\makecell[c]{Total}} \\
        ~ \\
        ~ \\
        ~ \\
        \midrule
        OpenFlamingo-9B \cite{alayrac2022flamingo, awadalla2023openflamingo} & 14.9 & 42.7 & 6.1 & 18.3 & 20.2 & 19.3 & 14.6 & 8.1 & 6.7 & 13.3 & 0.0 & 46.2 & 2.8 & 16.7 & 0.0 & 30.4 & 17.6$\pm$0.2 \\
        Otter-9B \cite{li2023otter} & 20.5 & 48.2 & 5.0 & 38.8 & 27.2 & 47.8 & 17.3 & 10.9 & 14.5 & 8.4 & 0.0 & 23.8 & 17.7 & 12.5 & 0.0 & 58.0 & 23.2$\pm$0.1 \\
        LLaVA-v1.5-7B \cite{llava} & 23.7 & 64.2 & 24.5 & 36.5 & 38.4 & 58.2 & 19.6 & 10.9 & 13.3 & 2.9 & 0.5 & 15.4 & 24.6 & 12.5 & 0.0 & 37.0 & 28.3$\pm$0.2\\
        LLaVA-v1.5-13B \cite{llava} & 26.5 & 60.7 & 33.9 & 55.5 & 49.9 & 43.2 & 28.5 & 12.8 & 13.3 & 16.9 & 0.0 & 26.9 & 43.4 & 20.8 & 0.0 & 44.6 & 33.2$\pm$0.1\\
        CogAgent-Chat \cite{cogagent} & 30.1 & 59.7 & 29.3 & 33.6 & 64.9 & 38.9 & 44.2 & 3.5 & 13.3 & 26.6 & 26.5 & 38.5 & 60.2 & 8.3 & 0.0 & 26.4 & 34.7$\pm$0.2 \\
        Emu2-Chat \cite{emu2} & 34.6 & 67.1 & 38.8 & 28.6 & 64.4 & 61.4 & 48.0 & 5.2 & 13.3 & 21.7 & 5.7 & 49.2 & 38.3 & 8.3 & 0.0 & 59.4 & 38.0$\pm$0.1 \\
        IXC2-VL-7B \cite{ixc2} & 40.9 & 66.9 & 40.4 & 60.9 & 71.1 & 64.6 & 48.1 & 20.4 & 46.0 & 3.6 & 13.8 & 15.4 & 56.3 & 0.0 & 36.4 & 30.8 & 42.5$\pm$0.3\\
        CogVLM-Chat \cite{cogvlm} & 44.4 & 72.2 & 36.9 & 46.1 & 53.6 & 59.6 & 55.7 & 18.7 & 10.8 & 32.7 & 22.2 & 37.7 & 61.2 & 33.3 & 9.1 & \thirdc{} 65.0 & 45.1$\pm$0.2\\
        InternVL-Chat-V1-2 \cite{internvl} & 39.6 & 69.9 & 46.9 & 57.9 & 75.7 & 71.8 & 48.1 & 22.2 & 13.3 & 32.3 & 18.6 & 29.2 & 64.2 & 16.7 & 9.1 & 35.2 & 45.5$\pm$0.1\\
        LLaVA-NeXT-34B \cite{llava} & 53.2 & 66.7 & 57.0 & 55.1 & 71.6 & 65.3 & 65.0 & 44.4 & 46.7 & 10.0 & 32.3 & 30.3 & 73.5 & 16.7 & 34.5 & 41.6 & 50.9$\pm$0.1\\
        InternVL-Chat-V1-5 \cite{internvl} & 47.5 & 72.1 & 59.2 & 56.6 & 69.1 & 65.3 & 69.2 & 37.7 & 23.6 & 42.7 & 22.9 & 43.1 & 70.8 & 25.0 & 0.0 & 43.4 & 51.5$\pm$0.2\\
        InternVL2-40B & 57.8 & \thirdc{} 77.7 & 70.0 & 75.0 & 81.6 & 75.0 & 62.3 & 59.9 & 53.3 & 57.9 & 54.6 & 38.9 & \secondc{} 86.5 & 25.0 & 40.9 & \secondc{} 66.8  & 63.8$\pm$0.2 \\
        InternVL2-Llama3-76B \cite{internvl2blog} & \secondc{} 62.4 & \firstc{} 80.4 & 74.5 & \secondc{} 77.7 & 75.2 & \firstc{} 78.1 & 79.5 & \secondc{} 60.5 & 66.7 & 67.7 & \thirdc{} 74.8 & 50.0 & \firstc{} 89.1 & 25.0 & \firstc{} 63.6 & 47.4 & \thirdc{} 68.4$\pm$0.3 \\
        \hdashline 
        Claude 3 Opus \cite{claude3} & 53.9 & 63.8 & 67.5 & 72.1 & 76.4 & 45.0 & 64.8 & 34.1 & 46.7 & 46.9 & 57.4 & 29.2 & 62.5 & 25.0 & 50.0 & 29.0 & 55.8$\pm$0.2 \\
        Qwen-VL-Max \cite{qwenvl} & 52.6 & 71.2 & 53.5 & 59.2 & 76.8 & 64.6 & 76.7 & 41.9 & 66.0 & 33.4 & 68.6 & \secondc{} 55.4 & 79.1 & 8.3 & \firstc{} 63.6 & 15.0 & 55.8$\pm$0.2 \\
        Gemini Pro Vision \cite{gemini} & 45.4 & 67.3 & 51.3 & 59.1 & 83.8 & \thirdc{} 76.1 & 76.7 & 48.6 & 52.0 & 39.9 & 50.6 & \firstc{} 74.6 & 67.4 & 33.3 & 51.8 & 40.0 & 57.2$\pm$0.2 \\
        GPT-4V \cite{gpt4v} & 56.4 & 71.0 & \firstc{} 82.4 & \thirdc{} 77.6 & \thirdc{} 85.5 & 63.1 & \thirdc{} 80.4 & \firstc{} 62.6 & \thirdc{} 86.0 & 62.7 & 64.2 & 37.7 & 80.0 & \secondc{} 41.7 & \firstc{} 63.6 & 35.0 & 66.3$\pm$0.2 \\
        Gemini 1.5 Pro \cite{gemini1_5} & 54.3 & 71.0 & 79.0 & 70.3 & \secondc{} 86.6 & \secondc{} 77.1 & 73.1 & 44.4 & 73.3 & \secondc{} 77.1 & \secondc{} 78.9 & \firstc{} 74.6 & 76.0 & \secondc{} 41.7 & \firstc{} 63.6 & 45.0 & 66.9$\pm$0.2 \\
        GPT-4o \cite{gpt4oblog} & \secondc{} 62.4 & 71.1 & \thirdc{} 79.1 & \firstc{} 79.5 & \firstc{} 89.9 & 65.5 & \secondc{} 82.5 & \thirdc{} 54.1 & \firstc{} 99.3 & \thirdc{} 75.7 & 73.5 & 46.8 & \thirdc{} 86.2 & 25.0 & \firstc{} 63.6 & \firstc{} 72.6 & \secondc{} 71.0$\pm$0.2 \\
        Claude 3.5 Sonnet \cite{claude3.5sonnet_blog} & \firstc{} 65.2 & \secondc{} 79.7 & \secondc{} 79.2 & 68.5 & 83.8 & 73.6 & \firstc{} 86.5 & 48.4 & \secondc{} 86.7 & \firstc{} 81.7 & \firstc{} 80.6 & 50.0 & 83.8 & \firstc{} 66.7 & 54.5 & 41.0 & \firstc{} 71.8$\pm$0.2 \\
        \bottomrule
    \end{tabular}
    }
\end{table}

\subsection{Experiment settings}
We evaluate two types of LMMs on our \benchnametwo{}: (1) open-weight LMMs including OpenFlamingo \cite{alayrac2022flamingo, awadalla2023openflamingo}, Otter \cite{li2023otter}, LLaVA \cite{llava, liu2024improved}, CogAgent \cite{cogagent}, Emu2-Chat \cite{emu2}, InternLM-XComposer2 (IXC2) \cite{ixc2}, CogVLM \cite{cogvlm} and InternVL \cite{internvl}; (2) Closed source models including Claude \cite{claude3, claude3.5sonnet_blog}, Qwen-VL Max \cite{qwenvl}, Gemini \cite{gemini, gemini1_5}, GPT-4V \cite{gpt4v} and GPT-4o \cite{gpt4oblog}.

As illustrated in \ref{tab:v2prompt}, for each sample, we complete the prompt template with its question, ground truth, and the output from a specific LMM. When this filled prompt is input into GPT-4, it generates a score ranging from 0 to 1 for each sample. Although the temperature is set to 0, we observe some variance in GPT-4's outputs. To address this, we evaluate the outputs of the LLMs using GPT-4 five times. Due to space constraints, we present the average scores for capabilities and for some capability integrations, and include both the average and variance for the total score.

\subsection{Results}
The main results of the different methods are presented in Table \ref{tab:results_cap} for each capability, and in Table \ref{tab:results_cap_integration} for 16 integrations with the highest proportions.
Claude 3.5 Sonnet \cite{claude3.5sonnet_blog} and GPT-4o \cite{gpt4oblog} are the leading models, achieving scores of 71.8 and 71.0, respectively. Claude 3.5 Sonnet \cite{claude3.5sonnet_blog} excels in recognition, language generation, OCR, spatial awareness, and knowledge. On the other hand, GPT-4o \cite{gpt4oblog} surpasses in image-text sequence understanding and math. Among open-weight models, InternVL2-Llama3-76B stands out with a competitive performance score of 68.4.

\section{Conclusion}
In this paper, we aim to evaluate the integrated capabilities of large multimodal models and extend MM-Vet into \benchnametwo{} by introducing a new core capability: Image-text sequence understanding, which assesses the ability to process vision-language sequences. Additionally, we ensure the high quality of evaluation samples while expanding the evaluation set size. Using \benchnametwo{} to benchmark large multimodal models, we found that Claude 3.5 Sonnet achieved the highest score of 71.8, narrowly surpassing GPT-4o, which scored 71.0. Among open-weight models, InternVL2-Llama3-76B emerged as the leader with a score of 68.4.

{
\small
\bibliographystyle{plain}
\bibliography{references}

\begin{thebibliography}{10}

\bibitem{claude3.5sonnet_blog}
Claude 3.5 sonnet.
\newblock \url{https://www.anthropic.com/news/claude-3-5-sonnet}, 2024.

\bibitem{gpt4oblog}
Hello gpt-4o.
\newblock \url{https://openai.com/index/hello-gpt-4o/}, 2024.

\bibitem{internvl2blog}
Internvl2: Better than the best—expanding performance boundaries of open-source multimodal models with the progressive scaling strategy.
\newblock \url{https://internvl.github.io/blog/2024-07-02-InternVL-2.0/}, 2024.

\bibitem{alayrac2022flamingo}
Jean-Baptiste Alayrac, Jeff Donahue, Pauline Luc, Antoine Miech, Iain Barr, Yana Hasson, Karel Lenc, Arthur Mensch, Katherine Millican, Malcolm Reynolds, et~al.
\newblock Flamingo: a visual language model for few-shot learning.
\newblock {\em Advances in Neural Information Processing Systems}, 35:23716--23736, 2022.

\bibitem{claude3}
Anthropic.
\newblock Claude 3 model card.
\newblock 2024.

\bibitem{awadalla2023openflamingo}
Anas Awadalla, Irena Gao, Josh Gardner, Jack Hessel, Yusuf Hanafy, Wanrong Zhu, Kalyani Marathe, Yonatan Bitton, Samir Gadre, Shiori Sagawa, Jenia Jitsev, Simon Kornblith, Pang~Wei Koh, Gabriel Ilharco, Mitchell Wortsman, and Ludwig Schmidt.
\newblock Openflamingo: An open-source framework for training large autoregressive vision-language models.
\newblock {\em arXiv preprint arXiv:2308.01390}, 2023.

\bibitem{qwenvl}
Jinze Bai, Shuai Bai, Shusheng Yang, Shijie Wang, Sinan Tan, Peng Wang, Junyang Lin, Chang Zhou, and Jingren Zhou.
\newblock Qwen-vl: A versatile vision-language model for understanding, localization, text reading, and beyond.
\newblock 2023.

\bibitem{internvl}
Zhe Chen, Jiannan Wu, Wenhai Wang, Weijie Su, Guo Chen, Sen Xing, Muyan Zhong, Qinglong Zhang, Xizhou Zhu, Lewei Lu, et~al.
\newblock Internvl: Scaling up vision foundation models and aligning for generic visual-linguistic tasks.
\newblock In {\em Proceedings of the IEEE/CVF Conference on Computer Vision and Pattern Recognition}, pages 24185--24198, 2024.

\bibitem{ixc2}
Xiaoyi Dong, Pan Zhang, Yuhang Zang, Yuhang Cao, Bin Wang, Linke Ouyang, Xilin Wei, Songyang Zhang, Haodong Duan, Maosong Cao, et~al.
\newblock Internlm-xcomposer2: Mastering free-form text-image composition and comprehension in vision-language large model.
\newblock {\em arXiv preprint arXiv:2401.16420}, 2024.

\bibitem{fu2023mme}
Chaoyou Fu, Peixian Chen, Yunhang Shen, Yulei Qin, Mengdan Zhang, Xu~Lin, Zhenyu Qiu, Wei Lin, Jinrui Yang, Xiawu Zheng, et~al.
\newblock Mme: A comprehensive evaluation benchmark for multimodal large language models.
\newblock {\em arXiv preprint arXiv:2306.13394}, 2023.

\bibitem{han2023chartllama}
Yucheng Han, Chi Zhang, Xin Chen, Xu~Yang, Zhibin Wang, Gang Yu, Bin Fu, and Hanwang Zhang.
\newblock Chartllama: A multimodal llm for chart understanding and generation.
\newblock {\em arXiv preprint arXiv:2311.16483}, 2023.

\bibitem{cogagent}
Wenyi Hong, Weihan Wang, Qingsong Lv, Jiazheng Xu, Wenmeng Yu, Junhui Ji, Yan Wang, Zihan Wang, Yuxiao Dong, Ming Ding, et~al.
\newblock Cogagent: A visual language model for gui agents.
\newblock In {\em Proceedings of the IEEE/CVF Conference on Computer Vision and Pattern Recognition}, pages 14281--14290, 2024.

\bibitem{li2023otter}
Bo~Li, Yuanhan Zhang, Liangyu Chen, Jinghao Wang, Jingkang Yang, and Ziwei Liu.
\newblock Otter: A multi-modal model with in-context instruction tuning.
\newblock {\em arXiv preprint arXiv:2305.03726}, 2023.

\bibitem{li2023seedbench}
Bohao Li, Rui Wang, Guangzhi Wang, Yuying Ge, Yixiao Ge, and Ying Shan.
\newblock Seed-bench: Benchmarking multimodal llms with generative comprehension, 2023.

\bibitem{lin2023mm}
Kevin Lin, Faisal Ahmed, Linjie Li, Chung-Ching Lin, Ehsan Azarnasab, Zhengyuan Yang, Jianfeng Wang, Lin Liang, Zicheng Liu, Yumao Lu, et~al.
\newblock Mm-vid: Advancing video understanding with gpt-4v (ision).
\newblock {\em arXiv preprint arXiv:2310.19773}, 2023.

\bibitem{liu2024improved}
Haotian Liu, Chunyuan Li, Yuheng Li, and Yong~Jae Lee.
\newblock Improved baselines with visual instruction tuning.
\newblock In {\em Proceedings of the IEEE/CVF Conference on Computer Vision and Pattern Recognition}, pages 26296--26306, 2024.

\bibitem{llava}
Haotian Liu, Chunyuan Li, Qingyang Wu, and Yong~Jae Lee.
\newblock Visual instruction tuning.
\newblock {\em arXiv preprint arXiv:2304.08485}, 2023.

\bibitem{liu2023mmbench}
Yuan Liu, Haodong Duan, Yuanhan Zhang, Bo~Li, Songyang Zhang, Wangbo Zhao, Yike Yuan, Jiaqi Wang, Conghui He, Ziwei Liu, et~al.
\newblock Mmbench: Is your multi-modal model an all-around player?
\newblock {\em arXiv preprint arXiv:2307.06281}, 2023.

\bibitem{openai2023gpt4}
OpenAI.
\newblock Gpt-4 technical report, 2023.

\bibitem{gpt4v}
OpenAI.
\newblock Gpt-4v(ision) system card.
\newblock 2023.

\bibitem{qinghong2024videogui}
Kevin Qinghong~Lin, Linjie Li, Difei Gao, Qinchen WU, Mingyi Yan, Zhengyuan Yang, Lijuan Wang, and Mike~Zheng Shou.
\newblock Videogui: A benchmark for gui automation from instructional videos.
\newblock {\em arXiv e-prints}, pages arXiv--2406, 2024.

\bibitem{gemini1_5}
Machel Reid, Nikolay Savinov, Denis Teplyashin, Dmitry Lepikhin, Timothy Lillicrap, Jean-baptiste Alayrac, Radu Soricut, Angeliki Lazaridou, Orhan Firat, Julian Schrittwieser, et~al.
\newblock Gemini 1.5: Unlocking multimodal understanding across millions of tokens of context.
\newblock {\em arXiv preprint arXiv:2403.05530}, 2024.

\bibitem{si2024design2code}
Chenglei Si, Yanzhe Zhang, Zhengyuan Yang, Ruibo Liu, and Diyi Yang.
\newblock Design2code: How far are we from automating front-end engineering?
\newblock {\em arXiv preprint arXiv:2403.03163}, 2024.

\bibitem{sun2023generative}
Quan Sun, Yufeng Cui, Xiaosong Zhang, Fan Zhang, Qiying Yu, Zhengxiong Luo, Yueze Wang, Yongming Rao, Jingjing Liu, Tiejun Huang, et~al.
\newblock Generative multimodal models are in-context learners.
\newblock {\em arXiv preprint arXiv:2312.13286}, 2023.

\bibitem{emu2}
Quan Sun, Yufeng Cui, Xiaosong Zhang, Fan Zhang, Qiying Yu, Yueze Wang, Yongming Rao, Jingjing Liu, Tiejun Huang, and Xinlong Wang.
\newblock Generative multimodal models are in-context learners.
\newblock In {\em Proceedings of the IEEE/CVF Conference on Computer Vision and Pattern Recognition}, pages 14398--14409, 2024.

\bibitem{gemini}
Gemini Team, Rohan Anil, Sebastian Borgeaud, Yonghui Wu, Jean-Baptiste Alayrac, Jiahui Yu, Radu Soricut, Johan Schalkwyk, Andrew~M Dai, Anja Hauth, et~al.
\newblock Gemini: a family of highly capable multimodal models.
\newblock {\em arXiv preprint arXiv:2312.11805}, 2023.

\bibitem{cogvlm}
Weihan Wang, Qingsong Lv, Wenmeng Yu, Wenyi Hong, Ji~Qi, Yan Wang, Junhui Ji, Zhuoyi Yang, Lei Zhao, Xixuan Song, et~al.
\newblock Cogvlm: Visual expert for pretrained language models.
\newblock {\em arXiv preprint arXiv:2311.03079}, 2023.

\bibitem{yan2023gpt}
An~Yan, Zhengyuan Yang, Wanrong Zhu, Kevin Lin, Linjie Li, Jianfeng Wang, Jianwei Yang, Yiwu Zhong, Julian McAuley, Jianfeng Gao, et~al.
\newblock Gpt-4v in wonderland: Large multimodal models for zero-shot smartphone gui navigation.
\newblock {\em arXiv preprint arXiv:2311.07562}, 2023.

\bibitem{yang2023dawn}
Zhengyuan Yang, Linjie Li, Kevin Lin, Jianfeng Wang, Chung-Ching Lin, Zicheng Liu, and Lijuan Wang.
\newblock The dawn of lmms: Preliminary explorations with gpt-4v (ision).
\newblock {\em arXiv preprint arXiv:2309.17421}, 2023.

\bibitem{you2024ferret}
Keen You, Haotian Zhang, Eldon Schoop, Floris Weers, Amanda Swearngin, Jeffrey Nichols, Yinfei Yang, and Zhe Gan.
\newblock Ferret-ui: Grounded mobile ui understanding with multimodal llms.
\newblock {\em arXiv preprint arXiv:2404.05719}, 2024.

\bibitem{yu2023mm}
Weihao Yu, Zhengyuan Yang, Linjie Li, Jianfeng Wang, Kevin Lin, Zicheng Liu, Xinchao Wang, and Lijuan Wang.
\newblock Mm-vet: Evaluating large multimodal models for integrated capabilities.
\newblock {\em arXiv preprint arXiv:2308.02490}, 2023.

\bibitem{yue2024mmmu}
Xiang Yue, Yuansheng Ni, Kai Zhang, Tianyu Zheng, Ruoqi Liu, Ge~Zhang, Samuel Stevens, Dongfu Jiang, Weiming Ren, Yuxuan Sun, et~al.
\newblock Mmmu: A massive multi-discipline multimodal understanding and reasoning benchmark for expert agi.
\newblock In {\em Proceedings of the IEEE/CVF Conference on Computer Vision and Pattern Recognition}, pages 9556--9567, 2024.

\bibitem{zhang2024mm}
Chaoyi Zhang, Kevin Lin, Zhengyuan Yang, Jianfeng Wang, Linjie Li, Chung-Ching Lin, Zicheng Liu, and Lijuan Wang.
\newblock Mm-narrator: Narrating long-form videos with multimodal in-context learning.
\newblock In {\em Proceedings of the IEEE/CVF Conference on Computer Vision and Pattern Recognition}, pages 13647--13657, 2024.

\end{thebibliography}
}

\end{document}